# FPGA Implementations of 3D-SIMD Processor Architecture for Deep Neural Networks Using Relative Indexed Compressed Sparse Filter Encoding Format and Stacked Filters Stationary Flow

Yuechao Gao[1, a)], Nianhong Liu[1, b)] and Sheng Zhang[1, c)]

[1]*Tsinghua National Laboratory for Information Science and Technology,
Institute of Microelectronics, Tsinghua University, Beijing 100084, China*
a) gyc15@mails.tsinghua.edu.cn
b) lnh15@mails.tsinghua.edu.cn
c) zhang_sh@tsinghua.edu.cn (Corresponding author.)

**Abstract.** It is a challenging task to deploy computationally and memory intensive State-of-the-art deep neural networks (DNNs) on embedded systems with limited hardware resources and power budgets. Recently developed techniques like Deep Compression make it possible to fit large DNNs, such as AlexNet and VGGNet, fully in on-chip SRAM. But sparse networks compressed using existing encoding formats, like CSR or CSC, complex the computation at runtime due to their irregular memory access characteristics. In [1], we introduce a computation dataflow, stacked filters stationary dataflow (SFS), and a corresponding data encoding format, relative indexed compressed sparse filter format (CSF), to make the best of data sparsity, and simplify data handling at execution time. In this paper we present FPGA implementations of these methods. We implement several compact streaming fully connected (FC) and Convolutional (CONV) neural network processors to show their efficiency. Comparing with the state-of-the-art results [2,3,4], our methods achieve at least $2\times$ improvement for computation efficiency per PE on most layers. Especially, our methods achieve $8\times$ improvement on AlexNet layer CONV4 with 384 filters, and $11\times$ improvement on VGG16 layer CONV5-3 with 512 filters.

## INTRODUCTION

In recent years, DNN technology has made breakthroughs in many areas such as motion detection, object detection, image classification and recognition, image semantic understanding, natural language processing, translation, and many other areas. A lot of groundbreaking neural networks have been proposed, such as AlexNet [11], VGG [12], GoogleNet [13], ResNet [14], R-FCN [15], Deformable-ConvNets [16], and so on. However, these networks contain millions of parameters, tens to hundreds of convolution layers, and require billions of arithmetic operations. They will also produce tons of intermediate data and need frequent data transmissions between process units and memory. The large amount of computation and huge memory resource consumption characteristics have hindered their wide usage in embedded devices. Various efforts have been made to address this issue, such as ShiftCNN [8], Ristretto [9], Eyeriss [4], Deep Compression [2] and EIE [3], etc. Among these approaches, deep compression is a promising method for embedded applications.

In our previous work [1], we pointed out several still existing problems of these approaches. The first problem is manipulating compressed sparse data need considerable extra logics and consumes extra clock cycles. Eyeriss [4] uses network on chip (NoC) to handle sparsity by only performing data reads and MACs on nonzero values; DVAS [5] and ENVISION [6] use input guard memories and guard control units to handle data sparsity. Several existing sparse matrix encoding formats, such as CSC, CSR and CISR [10], complex the computation at runtime due to their irregular memory access characteristics. This results in inefficiency in parallelizing computation and bigger chip area. The second one is, for deeply compressed sparse networks, the PE array utilization rate of recently proposed hardware acceleration designs, such as Eyeriss [4], DVAS [5], ENVISION [6], DNPU [7], etc., is fairly low. In [1], we proposed stacked filters stationary dataflow (SFS), relative indexed compressed sparse filter format (CSF), and a three dimensional Single Instruction Multiple Data (3D-SIMD) processor architecture to address these problems. Using these methods the sparse data can be easily handled during execution without complex transformations, lookups and computation. In this paper, we implement several compact streaming fully connected (FC) and Convolutional (CONV) neural network processors to show their efficiency. For the convenience of later usage, we copy equation (4) from [1] to equation (1) here.

$$V_{o'}^{(n)}[j][y][x] = \sum_{chi=0}^{C-1} \sum_{r=0}^{K-1} \sum_{c=0}^{K-1} W_{f'}^{(n)}[chi][r][c][j] \times V_i[chi][Sy+r][Sx+c] \quad (1)$$

$$0 \leq chi < C, 0 \leq r < K, 0 \leq c < K, 0 \leq j < m, 0 \leq x < W', 0 \leq y < H',$$
$$0 \leq n < M', M' = M/m, W' = (W-K)/S+1, H' = (H-K)/S+1. \quad (2)$$

$$V_{o'}^{(n)}[j][0][0] = \sum_{chi=0}^{C-1} \sum_{r=0}^{H-1} \sum_{c=0}^{W-1} W_{f'}^{(n)}[chi][r][c][j] \times V_i[chi][r][c] \quad (3)$$

$$W_{di} = (W_{do} - 1) \times S + K, H_{di} = (H_{do} - 1) \times S + K \quad (4)$$

Vo, Vi and Wf are the matrices of output feature maps, input feature maps and filters, respectively. S, C, K, M, M′, m, W, H, W′, H′ are a given stride size, channel number, filter kernel size, total filter number, number of batches, batch size, input feature width, height and output feature width, height.

## COMPUTATION FLOW OPTIMIZATION AND PARALLELIZATION

### FC Layer

For equation (1), one channel of feature data will convolute with m filters from the same channel in parallel. As the maximum output feature number of FC layers usually is 4096, all the output feature data can be buffered in the on-chip local registers. Grouping the filters is not needed, so to take full advantage of CSF and SFS, batch number M′=1 is used. As the output dimensions of FC layers are $M \times 1 \times 1$, computations of FC layers in DNNs can be simplified into equation (3). M is the number of filters in the current layer. The FC layer parallel computing pseudo code can be rewritten to the code as shown in figure 1. There are total $W \times H \times C$ number of feature values Vi and also the relative column pointer values. W, H and C are the input feature width, height and channel number, respectively. All the input values can be streamed into the processing units. The computation flow is: for each loop, first load one input feature value Vi[p], one pointer value ptr[p], ptr[p] number of filter weight values and their relative filter index values, accumulate the relative filter index to get the absolute index, and then Vi[p] multiplies with all the filter weight values just loaded and accumulate with their corresponding output feature values, which are decided by the absolute index. After looping $W \times H \times C$ times, the output feature values can be streamed out for next processing stage.

$$\text{for } p \text{ in } [0, W \times H \times C - 1] \{$$
$$\quad V_{o'}[ind[0]] += W_{f'}[0] \times V_i[p]$$
$$\quad V_{o'}[ind[1]] += W_{f'}[1] \times V_i[p]$$
$$\quad \ldots$$
$$\quad V_{o'}[ind[ptr[p]-1]] += W_{f'}[ptr[p]-1] \times V_i[p]$$
$$\}$$

**FIGURE 1.** FC layer parallel computing pseudo code.

### CONV Layer

For DNNs with large output feature dimensions, like VGG16, the output feature size of a CONV layer can be up to 12MB if 32 bit floating point numbers are used. It is hard to buffer all the intermediate computing results on-chip, which can be dealt with two skills, filter grouping and feature or image division. For designs using CSF data encoding format and SFS computing flow, filter grouping may cause performance degradation. Using feature division, as shown in figure 2, only one portion of the input feature is processed each time. The output buffer needed in figure 2 is only 1/4 of the original size. The advantage of using these two skills is that the on-chip buffer size is greatly reduced. The buffer can be implemented using registers instead of RAM, which can increase the processing bandwidth and increase parallelism. The disadvantage is that it will greatly increase the number of filter weight data or feature data loaded in each reference operation. Table 1 and 2 compares these two methods. It shows that comparing with filter grouping, feature division uses only one-half output feature buffer size and the total number of data loaded increases less. Besides, feature division won't lose computation efficiency and filters are usually compressed, so feature division is recommended for handling large CONV layers. For feature division, features are divided according to output feature dimension. If the output feature height and width of a division are Hdo and Wdo, equation (4) illustrates how to calculate the input feature height and width (Hdi and Wdi) of this division.

The CONV layer parallel computing pseudo code can be rewritten to the code as shown in figure 3. The computation flow of a single 3D-SIMD instruction is illustrated in figure 4. For each input channel, input feature data and filter weight data of this channel are buffered before computation. For one 3D-SIMD computation, there are total $K \times K$ input feature values and relative column pointer values participate in the calculation. The computation flow is: first load one input feature value Vi[p] and one pointer value ptr[p], load ptr[p] number of filter weight values and their relative filter index values, accumulate the relative filter index to get the absolute index. And then Vi[p] multiplies with all the filter weight values just loaded and accumulate with their corresponding output feature values, which are decided by the absolute index. For one 3D-SIMD computation, this flow loops $K \times K$ times. After a 3D-SIMD computation is finished, all the m output feature data in the on-chip registers are then moved to the global buffer. After looping all above flow $W' \times H' \times C$ times, the output feature values can be streamed out for next processing stage.

| V1,1 | V1,2 | ⋯ | V1,14 | V1,15 | V1,16 | ⋯ | V1,28 |
|---|---|---|---|---|---|---|---|
| V2,1 | V2,2 | ⋯ | V2,14 | V2,15 | V2,16 | ⋯ | V2,28 |
| … | … | … | … | … | … | … | … |
| V14,1 | V14,2 | ⋯ | V14,14 | V14,15 | V14,16 | ⋯ | V14,28 |

| V15,1 | V15,2 | ⋯ | V15,14 | V15,15 | V15,16 | ⋯ | V15,28 |
|---|---|---|---|---|---|---|---|
| V16,1 | V16,2 | ⋯ | V16,14 | V16,16 | V16,16 | ⋯ | V16,28 |
| … | … | … | … | … | … | … | … |
| V28,1 | V28,2 | ⋯ | V28,14 | V28,15 | V28,16 | ⋯ | V28,28 |

**FIGURE 2.** Illustration of $2 \times 2$ feature division of a certain channel.

$$
\begin{aligned}
&\text{for } chi \text{ in } [0, C-1] \{ \\
&\quad \text{buffer } V_i[chi], W_{f'}^{(n)}[chi], ptr \\
&\quad \text{for } y \text{ in } [0, H'-1], x \text{ in } [0, W'-1] \{ \\
&\quad\quad \text{for } p \text{ in } [0, K \times K - 1], r, c \text{ in } [0, K-1] \{ \\
&\quad\quad\quad V_{o'}^{(n)}[ind[0]][y][x] += W_{f'}^{(n)}[chi][r][c][0] \times V_i[chi][Sy+r][Sx+c] \\
&\quad\quad\quad V_{o'}^{(n)}[ind[1]][y][x] += W_{f'}^{(n)}[chi][r][c][1] \times V_i[chi][Sy+r][Sx+c] \\
&\quad\quad\quad \ldots \\
&\quad\quad\quad V_{o'}^{(n)}[ind[ptr[p]-1]][y][x] += W_{f'}^{(n)}[chi][r][c][ptr[p]-1] \times V_i[chi][Sy+r][Sx+c] \\
&\quad\quad \} \\
&\quad \} \\
&\}
\end{aligned}
$$

**FIGURE 3.** CONV layer parallel computing pseudo code.

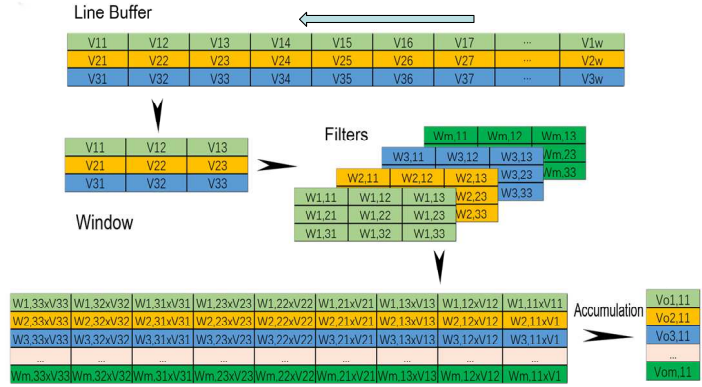

**FIGURE 4.** Illustration of a single 3D-SIMD computation flow.

**TABLE 1.** Feature division on VGG16 (output buffer size: 100352).

| Layer # | Feature division | Load times | Filter weight # | Total weight # loaded |
|---|---|---|---|---|
| CONV1-1 | $16 \times 16$ | 256 | 1728 | 442368 |
| CONV1-2 | $16 \times 16$ | 256 | 36864 | 9437184 |
| CONV2-1 | $8 \times 8$ | 64 | 73728 | 4718592 |
| CONV2-2 | $8 \times 8$ | 64 | 147456 | 9437184 |
| CONV3-1 | $4 \times 4$ | 16 | 294912 | 4718592 |
| CONV3-2 | $4 \times 4$ | 16 | 589824 | 9437184 |
| CONV3-3 | $4 \times 4$ | 16 | 589824 | 9437184 |
| CONV4-1 | $2 \times 2$ | 4 | 1179648 | 4718592 |
| CONV4-2 | $2 \times 2$ | 4 | 2359296 | 9437184 |
| CONV4-3 | $2 \times 2$ | 4 | 2359296 | 9437184 |
| CONV5-1 | $1 \times 1$ | 1 | 2359296 | 2359296 |
| CONV5-2 | $1 \times 1$ | 1 | 2359296 | 2359296 |
| CONV5-3 | $1 \times 1$ | 1 | 2359296 | 2359296 |
| Total | | | 14,710,464 | 78,299,136 |

**TABLE 2.** Filter grouping on VGG16 (output buffer size: 200704).

| Layer # | Filter batch size | Filter batches | Feature value # | Total feature # loaded |
|---|---|---|---|---|
| CONV1-1 | 4 | 16 | 150528 | 2408448 |
| CONV1-2 | 4 | 16 | 3211264 | 51380224 |
| CONV2-1 | 16 | 8 | 802816 | 6422528 |
| CONV2-2 | 16 | 8 | 1605632 | 12845056 |
| CONV3-1 | 64 | 4 | 401408 | 1605632 |
| CONV3-2 | 64 | 4 | 802816 | 3211264 |
| CONV3-3 | 64 | 4 | 802816 | 3211264 |
| CONV4-1 | 256 | 1 | 200704 | 200704 |
| CONV4-2 | 256 | 2 | 401408 | 802816 |
| CONV4-3 | 256 | 2 | 401408 | 802816 |
| CONV5-1 | 512 | 1 | 100352 | 100352 |
| CONV5-2 | 512 | 1 | 100352 | 100352 |
| CONV5-3 | 512 | 1 | 100352 | 100352 |
| Total | | | 9,081,856 | 83,191,808 |

# HARDWARE IMPLEMENTATION

Figure 5 shows the 3D-SIMD processor architecture. To lower computation complexity, skills described in ShiftCNN [8] are used to simplify floating point number multiplication. So the PEs in this paper only process 32 bit float point number shifts and additions. For FC layers, feature values, pointers, filter weight values and indices can all be streamed in from outside RAM, there is no need to buffer them internally if the data load speed can match the PE processing speed. For CONV layer, before processing a channel, input feature data and filter data are buffered into the on-chip global buffer first. All the designs are implemented on a Xilinx ZCU102 evaluation kit. This kit features a Zynq MPSoC device with ARM processors and programmable logic fabric. Figure 6 shows the resources used by a simple high speed streaming FC layer processor for LeNet. Figure 7 shows the resources used by a CONV layer processor for AlexNet. Figure 8 shows the resources used by the CONV layer processor for VGG16.

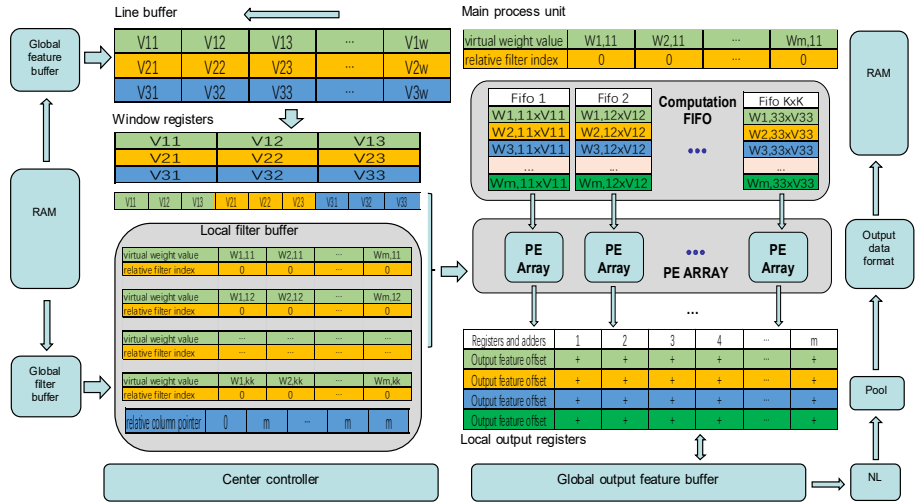

**FIGURE 5.** 3D-SIMD processor architecture.

**FIGURE 6.** LeNet FC processor.   **FIGURE 7.** AlexNet CONV processor.   **FIGURE 8.** VGG16 CONV processor.

# RESULT

The LeNet implementation is tested on the MNIST data set [17]. By using 8 PEs, this simple design can process this data set at 70 fps (Table 3). AlexNet and VGG16 implementations are all tested on ImageNet 2012 data set [18]. For AlexNet CONV processor, the first CONV layer of AlexNet is divided into 4×4 divisions. For VGG16 CONV processor, the first CONV layer of VGG16 is divided into 16×16 divisions. Table 4 and 5 illustrate their performance.

**TABLE 3.** A simple LeNet implementation.

| Layer # | Bits per operand | PE # | Clock Frequency (MHz) | # of MACs (millions of MACs) | Run Time (msec) |
|---|---|---|---|---|---|
| CONV1 | 32 | 8 | 299.97 | 0.3 | 1.7640 |
| CONV2 | 32 | 8 | 299.97 | 1.6 | 10.9350 |
| FC1 | 32 | 8 | 299.97 | 0.40 | 1.5000 |
| FC2 | 32 | 8 | 299.97 | 0.01 | 0.1170 |

TABLE 4. AlexNet CONV layer efficiency comparing with Eyeriss[4].

| Layer # | Bits per operand | PE # | Clock Frequency (MHz) | # of MACs (millions of MACs) | Run Time (msec) | Computation efficiency Per PE | Improvement (×) |
|---|---|---|---|---|---|---|---|
| CONV1(Our) | 32 | 8 | 299.97 | 105.4152 | 441.9150 | 0.0596 | 0.5 |
| CONV1[4] | 16 | 168 | 100-250 | 421.6608 | 20.9000 | 0.1201 | |
| CONV2(Our) | 32 | 8 | 299.97 | 447.8976 | 498.9550 | 0.2244 | 2.2 |
| CONV2[4] | 16 | 168 | 100-250 | 716.4638 | 41.9000 | 0.1018 | |
| CONV3(Our) | 32 | 8 | 299.97 | 149.5204 | 156.2320 | 0.2393 | 3.9 |
| CONV3[4] | 16 | 168 | 100-250 | 241.1512 | 23.6000 | 0.0608 | |
| CONV4(Our) | 32 | 8 | 299.97 | 224.2806 | 182.2490 | 0.3077 | 8.0 |
| CONV4[4] | 16 | 168 | 100-250 | 118.1646 | 18.4000 | 0.0382 | |
| CONV5(Our) | 32 | 8 | 299.97 | 149.5204 | 155.6620 | 0.3077 | 5.2 |
| CONV5[4] | 16 | 168 | 100-250 | 81.6753 | 10.5000 | 0.0463 | |

TABLE 5. VGG16 CONV layer efficiency comparing with Eyeriss[4].

| Layer # | Bits per operand | PE # | Clock Frequency (MHz) | # of MACs (millions of MACs) | Run Time (msec) | Computation efficiency Per PE | Improvement (×) |
|---|---|---|---|---|---|---|---|
| CONV1-1(our) | 32 | 8 | 299.97 | 86.7 | 605.9 | 0.0358 | 1.8 |
| CONV1-1[4] | 16 | 168 | 100-250 | 258.5 | 76.2 | 0.0202 | |
| CONV1-2(our) | 32 | 8 | 299.97 | 1849.7 | 6531.1 | 0.0708 | 3.7 |
| CONV1-2[4] | 16 | 168 | 100-250 | 2910.2 | 910.3 | 0.0190 | |
| CONV2-1(our) | 32 | 8 | 299.97 | 924.8 | 3678.0 | 0.0629 | 2.3 |
| CONV2-1[4] | 16 | 168 | 100-250 | 2133.0 | 470.3 | 0.0270 | |
| CONV2-2(our) | 32 | 8 | 299.97 | 1849.7 | 6014.1 | 0.0769 | 3.4 |
| CONV2-2[4] | 16 | 168 | 100-250 | 3371.8 | 894.3 | 0.0224 | |
| CONV3-1(our) | 32 | 8 | 299.97 | 924.8 | 2655.4 | 0.0871 | 2.1 |
| CONV3-1[4] | 16 | 168 | 100-250 | 1660.6 | 241.1 | 0.0410 | |
| CONV3-2(our) | 32 | 8 | 299.97 | 1849.7 | 4141.1 | 0.1117 | 3.4 |
| CONV3-2[4] | 16 | 168 | 100-250 | 2538.8 | 460.9 | 0.0328 | |
| CONV3-3(our) | 32 | 8 | 299.97 | 1849.7 | 4091.1 | 0.1130 | 3.7 |
| CONV3-3[4] | 16 | 168 | 100-250 | 2323.9 | 457.7 | 0.0302 | |
| CONV4-1(our) | 32 | 8 | 299.97 | 924.8 | 1977.2 | 0.1169 | 2.4 |
| CONV4-1[4] | 16 | 168 | 100-250 | 1109.0 | 135.8 | 0.0486 | |
| CONV4-2(our) | 32 | 8 | 299.97 | 1849.7 | 2689.9 | 0.1719 | 4.9 |
| CONV4-2[4] | 16 | 168 | 100-250 | 1503.0 | 254.8 | 0.0351 | |
| CONV4-3(our) | 32 | 8 | 299.97 | 1849.7 | 2586.6 | 0.1788 | 7.6 |
| CONV4-3[4] | 16 | 168 | 100-250 | 973.4 | 246.3 | 0.0235 | |
| CONV5-1(our) | 32 | 8 | 299.97 | 462.4 | 596.5 | 0.1938 | 5.3 |
| CONV5-1[4] | 16 | 168 | 100-250 | 333.3 | 54.3 | 0.0365 | |
| CONV5-2(our) | 32 | 8 | 299.97 | 462.4 | 548.9 | 0.2106 | 8.7 |
| CONV5-2[4] | 16 | 168 | 100-250 | 218.4 | 53.7 | 0.0242 | |
| CONV5-3(our) | 32 | 8 | 299.97 | 462.4 | 479.8 | 0.2410 | 11.0 |
| CONV5-3[4] | 16 | 168 | 100-250 | 198.5 | 53.7 | 0.0220 | |

# CONCLUSION

In this paper we present FPGA implementations of the proposed 3D-SIMD processor architecture. We implement several compact streaming FC and Convolutional neural network processors to show their efficiency. Comparing with the state-of-the-art result [2][3][4], our methods achieve at least 2× improvement for the computation efficiency per

PE on most layers. Especially, our methods achieve 8× improvement for the computation efficiency per PE on AlexNet layer CONV4 with 384 filters, and 11× improvement on VGG16 layer CONV5-3 with 512 filters.

For the future ASIC implementation of the 3D-SIMD processor architecture, approximating of the networks with 16 bit fixed point numbers should be considered to lower the computation complexity. All the experiments in this paper are done with 32-bit floating point numbers, which consume 11 clocks for one single addition computation. To have the best performance of CONV and FC layers, care should be taken for handling the filter parameters. If the maximum number of filter parameters can be loaded in one clock matches the maximum number of PEs used in the implementation, the best performance can be achieved.

## REFERENCES

1. Yuechao Gao, Nianhong Liu, and S. Zhang. Relative Indexed Compressed Sparse Filter Encoding Format for Hardware-Oriented Acceleration of Deep Convolutional Neural Networks. ISNE2018 (unpublished).
2. Song Han, Huizi Mao, and William J Dally. Deep compression: Compressing deep neural networks with pruning, trained quantization and huffman coding. International Conference on Learning Representations (ICLR), 2016b.
3. Song Han, Xingyu Liu, Huizi Mao, Jing Pu, Ardavan Pedram, Mark A Horowitz, and William J Dally. Eie: Efficient inference engine on compressed deep neural network. International Conference on Computer Architecture (ISCA), 2016a.
4. Yu Hsin Chen, Tushar Krishna, Joel S. Emer, and Vivienne Sze. Eyeriss: An energy-efficient reconfigurable accelerator for deep convolutional neural networks. IEEE Journal of Solid-State Circuits, 52(1):127–138, 2017.
5. Bert Moons and Marian Verhelst. Dvas: Dynamic voltage accuracy scaling for increased energy efficiency in approximate computing. In Ieee/acm International Symposium on Low Power Electronics and Design, pp. 237–242, 2015.
6. Bert Moons, Roel Uytterhoeven, Wim Dehaene, and Marian Verhelst. 14.5 envision: A 0.26-to-10tops/w subword-parallel dynamic-voltage-accuracy-frequency-scalable convolutional neural network processor in 28nm fdsoi. In Solid-State Circuits Conference, pp. 246–247, 2017.
7. Dongjoo Shin, Jinmook Lee, Jinsu Lee, and Hoi Jun Yoo. 14.2 dnpu: An 8.1tops/w reconfigurable cnn-rnn processor for general-purpose deep neural networks. In Solid-State Circuits Conference, pp. 240–241, 2017.
8. Denis A Gudovskiy and Luca Rigazio. Shiftcnn: Generalized low-precision architecture for inference of convolutional neural networks. 2017.
9. Philipp Gysel. Ristretto: Hardware-oriented approximation of convolutional neural networks. 2016.
10. Jeremy Fowers, Kalin Ovtcharov, Karin Strauss, Eric S. Chung, and Greg Stitt. A high memory bandwidth fpga accelerator for sparse matrix-vector multiplication. In IEEE International Symposium on Field-Programmable Custom Computing Machines, pp. 36–43, 2014.
11. Krizhevsky, Alex, I. Sutskever, and G. E. Hinton. ImageNet classification with deep convolutional neural networks. International Conference on Neural Information Processing Systems Curran Associates Inc. 2012:1097-1105.
12. K. Simonyan and A. Zisserman. Very deep convolutional networks for large-scale image recognition. In ICLR, 2015.
13. C. Szegedy, W. Liu, Y. Jia, P. Sermanet, S. Reed, D. Anguelov, D. Erhan, V. Vanhoucke, and A. Rabinovich. Going deeper with convolutions. In Proceedings of the IEEE Conference on Computer Vision and Pattern Recognition, pages 1–9, 2015.
14. K. He, X. Zhang, S. Ren, and J. Sun. Deep residual learning for image recognition. arXiv preprint arXiv:1512.03385, 2015.
15. Dai J, Li Y, He K, et al. R-FCN: Object Detection via Region-based Fully Convolutional Networks [J]. 2016.
16. Dai J, Qi H, Xiong Y, et al. Deformable Convolutional Networks [J]. 2017:764-773.
17. Y. LeCun, L. Bottou, Y. Bengio, and P. Haffner. Gradient-based learning applied to document recognition. Proc. IEEE, vol. 86, no. 11, pp. 2278–2324, Nov. 1998.
18. A. Krizhevsky, I. Sutskever, and G. E. Hinton. ImageNet classification with deep convolutional neural networks. In Proc. Adv. Neural Inf. Process. Syst., vol. 25. 2012, pp. 1097–1105.